\documentclass[10pt, twocolumn]{article}   	% use "amsart" instead of "article" for AMSLaTeX format
\usepackage{geometry}                		% See geometry.pdf to learn the layout options. There are lots.
\pdfoutput=1                 		% ... or a4paper or a5paper or ... 
\usepackage{graphicx}
\usepackage{subfigure}				% Use pdf, png, jpg, or eps with pdflatex; use eps in DVI mode
\usepackage{amssymb}
\usepackage{amsmath}
\usepackage{url}
\usepackage{hyperref}
\title{\bf Comparing Multilayer Perceptron and Multiple Regression Models for Predicting\\ Energy Use in the Balkans}
\author{Radmila Jankovi\'c$^1$, Alessia Amelio$^2$}
\date{\small $^1$Mathematical Institute of the S.A.S.A, Belgrade, Serbia, \href{mailto:rjankovic@mi.sanu.ac.rs}{\underline{rjankovic@mi.sanu.ac.rs}}\\$^2$DIMES, University of Calabria, Rende, Italy, \href{mailto:aamelio@dimes.unical.it}{\underline{aamelio@dimes.unical.it}}	}						

\begin{document}
%\twocolumn[
\maketitle

%\begin{@twocolumnfalse}
{\small \bf \textit{Abstract} -- Global demographic and economic changes have a critical impact on the total energy consumption, which is why demographic and economic parameters have to be taken into account when making predictions about the energy consumption. This research is based on the application of a multiple linear regression model and a neural network model, in particular multilayer perceptron, for predicting the energy consumption. Data from five Balkan countries has been considered in the analysis for the period 1995-2014. Gross domestic product, total number of population, and CO$_2$ emission were taken as predictor variables, while the energy consumption was used as the dependent variable. The analyses showed that CO$_2$ emissions have the highest impact on the energy consumption, followed by the gross domestic product, while the population number has the lowest impact. The results from both analyses are then used for making predictions on the same data, after which the obtained values were compared with the real values. It was observed that the multilayer perceptron model predicts better the energy consumption than the regression model.\\\\
{\bf Keywords} -- energy use, regression, neural network.}
%\end{@twocolumnfalse}
%\vspace{1cm}
%]

\section{Introduction}
The energy consumption worldwide is rising as the population numbers grow. It is estimated that the world population growth was 1.2\% in 2016 \cite{[1]}. Higher population numbers indicate a fast expansion of residential and industrial buildings, which further increases the energy consumption. The industrial processes also impact on the energy use, since the countries try to stay within positive economic growth rates. As the society becomes more demanding, the need to successfully manage the energy consumption rises. 

Usually, the modelling of the energy use is based on time series data that is often incomplete and complex. Creating a model that would be able to accurately predict the energy use is still challenging, as these models can be very complex and hard to manage \cite{[2]}. But, even when complex, these models are necessary for future planning and production of energy. 

A report by the Balkan green foundation \cite{[3]} indicated that in 2015, Serbia had the highest final consumption of electricity (28,551 GWh), compared to Bosnia and Herzegovina (11,183 GWh) and Macedonia (6,600 GWh). The consumption structure of these three countries indicates that more than 50\% of energy was used in the industrial sector, transport, services and agriculture, with the rest of the energy being used in households. Slovenia has obligated to improve its energy efficiency until 2020 by 20\% \cite{[4]}, while projections for Serbia state that the final energy consumption will increase of 10.1\% by 2020, compared to the energy consumption in 2010 \cite{[5]}. In the action plan it was reported that Serbia managed to accomplish savings of 4.4\% in the period from 2010 to 2015, with the goal of achieving 9\% savings in 2018 \cite{[6]}.  Croatia projected that the final energy consumption will increase of 5.85\% in 2020, compared to 2015, but it will decrease of 3.19\% compared to 2010 \cite{[7]}. Bosnia and Herzegovina set a reduction of 9\% in final energy consumption as a goal by the end of 2018 \cite{[8]}. All this clearly emphasizes the need for accurate and simple to use prediction models and techniques, as energy efficiency goals are hard to maintain if the energy forecast is not reliable. 

The aim of this paper is to investigate and model the energy consumption in West Balkan using two techniques: (i) multiple linear regression, and (ii) artificial neural network (ANN), in particular multilayer perceptron. The relationship between the energy use as a dependent variable and gross domestic product (GDP), population, and CO$_2$ emissions as independent variables was investigated. The purpose is to determine which variable has the biggest impact on the energy use, and to see which technique best predicts the values of energy consumption. Moreover, based on the regression analysis, the energy consumption elasticity was also calculated.
	
\section{Literature Review}
Kankal et al. \cite{[9]} modeled the energy consumption based on GDP, population, import, export, and employment as variables. They have developed different models based on a multiple linear regression analysis, power regression analysis, and ANN. It was concluded that the ANN model predicts the energy use better than the regression models. Borozan \cite{[10]} investigated the relationship between energy consumption (in particular liquid fuel, natural gas, hydropower, electricity and coal) and GDP in Croatia using the bivariate vector autoregression and Granger causality tests. It was concluded that savings in energy consumption can negatively affect the economic growth. Moreover, a causal relationship was not found between GDP and coal, but a relationship exists for other investigated forms of energy. In \cite{[11]}, the authors have compared regression models, neural networks and least squares support vector machine models for forecasting the electricity consumption. As predictors in their study, gross electricity generation, installed capacity, total subscribership and population were used. The results indicate that the least squares support vector machine was the most effective for predicting electricity consumption. In \cite{[12]}, the authors applied an ANN model to monthly build electric energy consumption data. Their results indicated that the ANN model relatively accurately predicted the energy consumption, but the accuracy slightly varies depending on the prediction period. In \cite{[13]}, the authors developed a weighted multi Support Vector Regression (SVR) model with a differential evolution optimisation algorithm for predicting the buildingsÕ energy consumption. It was concluded that the model which used the optimization algorithm achieved higher accuracy and could be used for predicting daily and half-hourly energy consumption of a building. A decision tree model, in particular the Random Forest, and the ANN model were compared in \cite{[14]} in terms of predicting the energy consumption of a building. Based on the analysis, the authors concluded that ANN performed better than the Random Forest algorithm, with higher accuracy on the testing set of data.

Differently from the previous literature, this paper introduces a comparison between a regression model and an ANN model developed on a dataset from Balkan countries. We try to answer the following questions: (1) Can the energy use be predicted based on the values of CO$_2$ emissions, population, and GDP from five Balkan countries, and (2) What model better predicts the energy use in this situation. 

The paper is organised as follows. Section III describes the data and methodology used in the research. Section IV shows the results and makes a discussion. Finally, Section V draws conclusions from this research.  

\section{Data and methodology}
The purpose of this research was to investigate if the energy use can be successfully modelled by two techniques: (1) multiple linear regression, and (2) multilayer perceptron neural network. Multiple linear regression is a method used when there is one outcome and more than one predictor variable. It creates a statistical relation between the predictor variables and the outcome, and can be represented by the following equation \cite{[15]}:
\begin{equation}
Yi=\beta_0+\beta_1 x_{i1}+\beta_2 x_{i2}+...+\beta_{(p-1)} x_{(i,p-1)}+\epsilon_i 
\end{equation}
where $Yi$ is the dependent variable, $\beta_0$ is the intercept, $\beta_1$ is the coefficient of the independent variable, $x_i$ is the predictor variable, and $\epsilon_i$ is the random error.

The multilayer perceptron, on the other hand, is a type of ANN and consists of one or more input layers, hidden layers that are formed by nodes, and output layers \cite{[16]}. A linear activation function is contained in the neurons of the output layer, while in the hidden layer this function is nonlinear. The nodes in the network are interconnected and actually represent units for information processing that are arranged in layers \cite{[16]}. The task of each type of layer differs, so the input layer receives the information, the hidden layer processes it, while the output layer performs predictions. Before passing information to the hidden layer, the values from the input layer are multiplied by weights, with results being added in order to get only one number. This number represents an argument to the activation function \cite{[17]}.  

The data used in this research has been downloaded from World Bank \cite{[1]}. The created dataset consists of six variables: (1) Country, (2) Year, (3) CO$_2$ emissions in metric tons per capita, (4) GDP per capita in constant 2010 US\$, (5) Energy use in kilograms of oil equivalent per capita, and (6) Total population. The research involved annual data from five Balkan countries: Bosnia and Herzegovina, Croatia, Former Yugoslav Republic of Macedonia, Serbia, and Slovenia, and included the period from 1995 to 2014. All analyses were performed using the software SPSS v.23 and Microsoft Excel 2010.

\begin{figure}[htbp]
\begin{center}
\includegraphics[height=7.2cm, width=7.2cm, keepaspectratio]{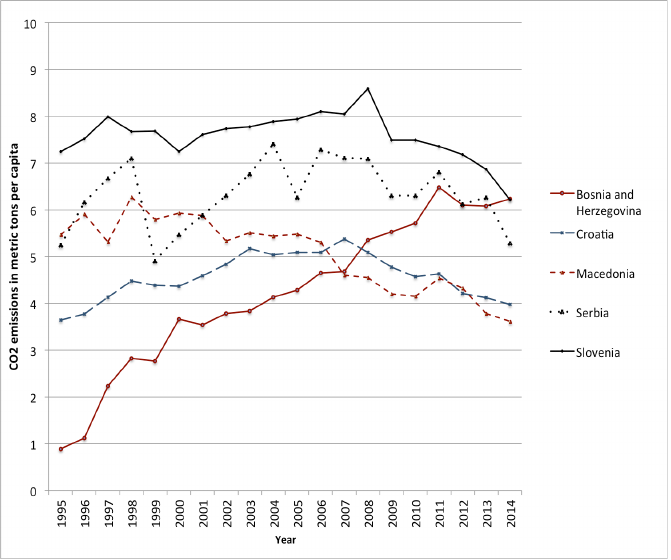}
\caption{CO$_2$ emissions in the observed period}
\label{fig1}
\end{center}
\end{figure}

The trend of CO$_2$ emissions in the observed period represented in metric tons per capita is shown in Fig. \ref{fig1}. The highest CO$_2$ emissions are observed for Slovenia, with a peak in 2008, while the lowest values are observed for Croatia and Macedonia. 

\begin{figure}[htbp]
\begin{center}
\includegraphics[height=7.2cm, width=7.2cm, keepaspectratio]{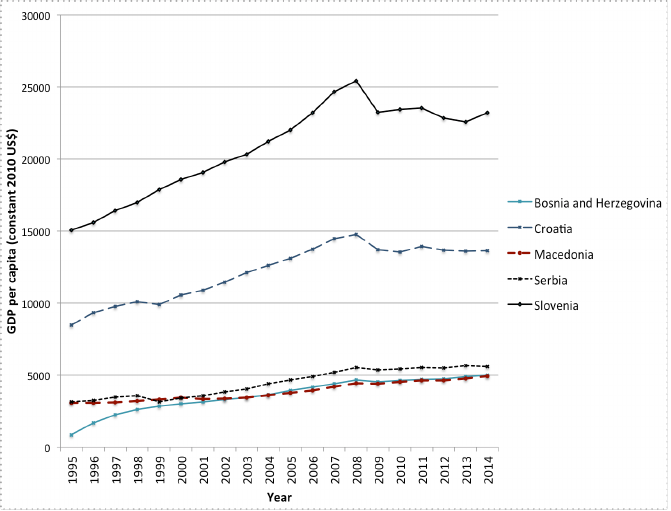}
\caption{GDP per capita for the observed period}
\label{fig2}
\end{center}
\end{figure}

Fig. \ref{fig2} shows the trend of GDP per capita for the period from 1994 to 2014. The highest GDP per capita is observed for Slovenia, while the lowest GDP per capita is found for Macedonia and Bosnia and Herzegovina. 

\begin{figure}[htbp]
\begin{center}
\includegraphics[height=7.2cm, width=7.2cm, keepaspectratio]{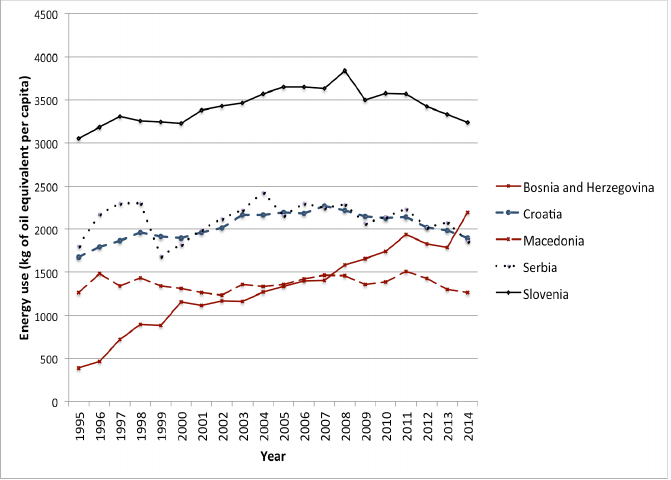}
\caption{Energy use in the observed period}
\label{fig3}
\end{center}
\end{figure}

The energy use in the observed period for the five Balkan countries is shown in Fig. \ref{fig3}. Based on the chart, the highest energy use is observed for Slovenia, while the lowest energy use is found for Macedonia. 

\begin{figure}[htbp]
\begin{center}
\includegraphics[height=7.2cm, width=7.2cm, keepaspectratio]{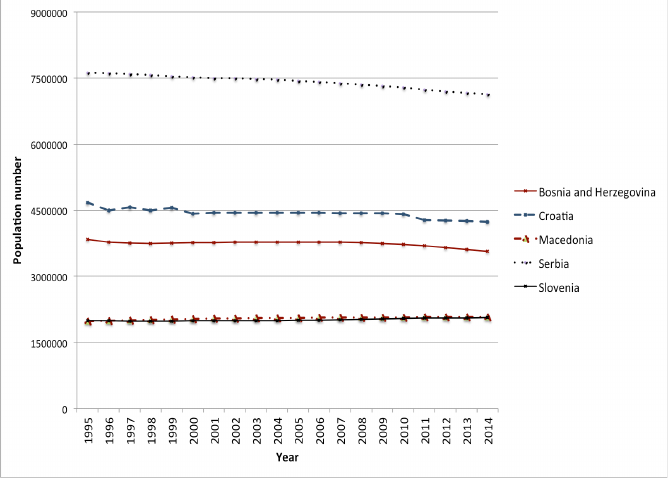}
\caption{Population of the tested countries}
\label{fig4}
\end{center}
\end{figure}

Considering the population of the tested countries, it is worth noting that the highest number of population is in Serbia, while Macedonia and Slovenia have the lowest number of population. These trends can be clearly observed from Fig. \ref{fig4}.  

In Table \ref{tab1}, descriptive statistics of the dataset -- mean, standard deviation, minimum and maximum values, can be observed for the five countries in terms of the following variables: (a) CO$_2$ emissions, (b) GDP, and (c) Energy use. As observed, the highest mean value of CO$_2$ emissions in metric tons per capita in the period from 1995 to 2014 can be found for Slovenia (M=7.58, SD=0.502), while the lowest is observed for Bosnia and Herzegovina (M=4.195, SD=1.647). Considering the GDP per capita, the highest mean can be found for Slovenia (M=20761.686, SD=3153.642), and the lowest one is found for Bosnia and Herzegovina (M=3617.605, SD=1157.083). The highest energy use can be observed for Slovenia (M=3424.976, SD=198.278), while the lowest energy consumers in this period are Bosnia and Herzegovina (M=1304.608, SD=484.574), followed by Macedonia (M=1365.916, SD=81.077).  
\begin{table*}[ht]
\caption{Descriptive statistics for the tested countries}
\begin{center}
\begin{tabular}{cc}
\includegraphics[height=11cm, width=11cm, keepaspectratio]{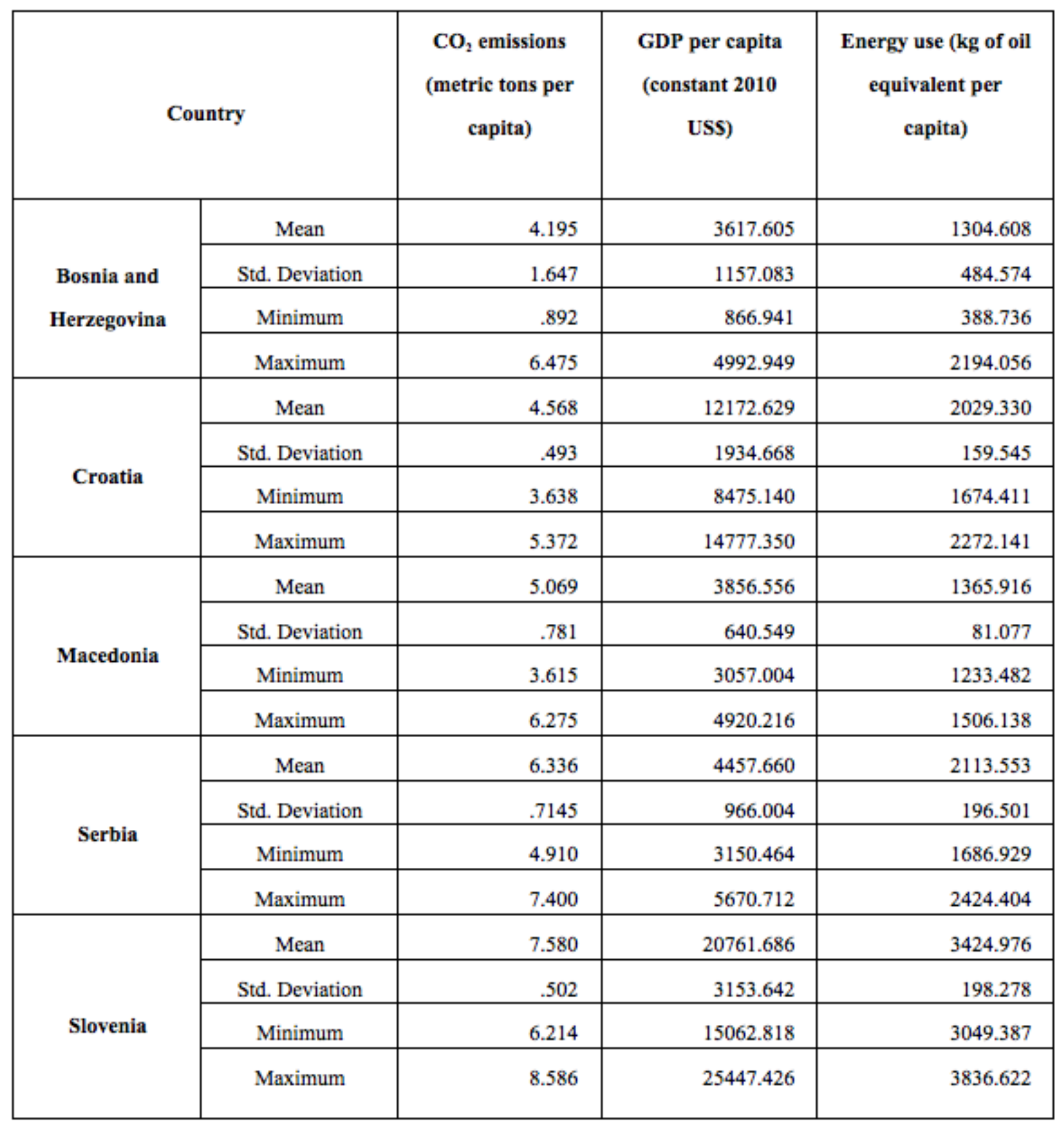}
\end{tabular}
\end{center}
\label{tab1}
\end{table*}%

\section{Analysis and discussion}
The analysis included the development of a multiple regression model and also a neural network model. These two models were then compared in terms of their root mean square error (RMSE) and mean absolute percentage error (MAPE).

\subsection{Multiple linear regression model}
The multiple regression model includes one dependent variable (Energy use) and three independent variables: (1) CO$_2$ emissions (metric tons per capita), (2) GDP per capita (constant 2010 US\$), and (3) Population. All variables were logarithmically transformed, and a log-log model was developed. A Stepwise regression method was used for developing a regression model, with the following criteria: probability of F to enter $\le$.050; probability of F to remove $\ge$.100. The F-statistic is used for testing the significance of the regression coefficients of the independent variables that will be included or excluded from the model. The regression equation for log-transformed models can be expressed as in Eq. (\ref{eq2}):
\begin{equation}\label{eq2}
ln(Y)=\beta_0+\beta_1ln(X_1)+\beta_2ln(X_2 )+\epsilon_i     
\end{equation}     

By applying the general equation to the data, the following regression equation can be generated:
\begin{equation}\label{eq3}
ln(EU)=\beta_0+\beta_1ln(GDP)+\beta_2ln(P)+\beta_3ln(CO_2)
\end{equation}

 where $EU$ represents the energy use, $GDP$ is the gross domestic product, $P$ is the population, and $CO_2$ are the emissions. 

Based on the results from the regression analysis (Table \ref{tab2}), it is visible that the energy use can be explained by GDP, CO$_2$ emissions and total population number with high fit ($R^2$=0.969), with a significant F-statistics value ($p$=0.000). 
\begin{table}[ht]
\caption{Results of multiple linear regression}
\begin{center}
\begin{tabular}{cc}
\includegraphics[height=7cm, width=7cm, keepaspectratio]{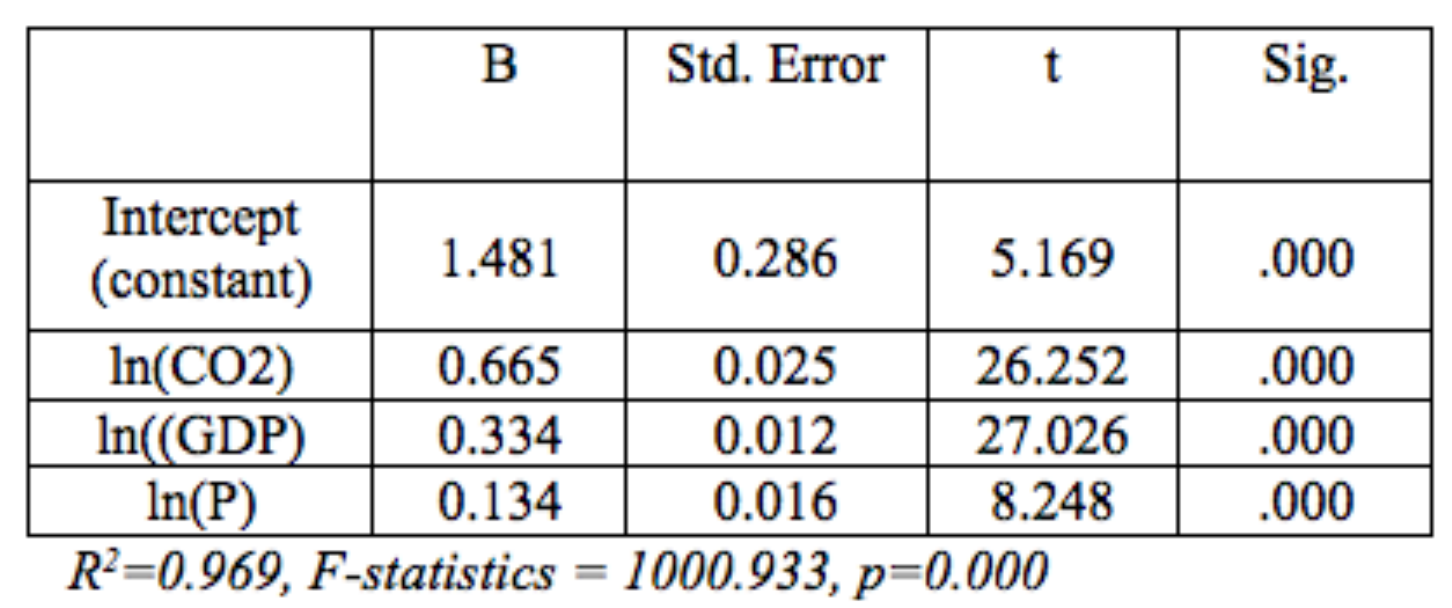}
\end{tabular}
\end{center}
\label{tab2}
\end{table}%  
 
By incorporating the regression results into Eq. (\ref{eq3}), the following regression equation can be generated:
\begin{multline}\label{eq4}
ln(\textrm{Energy use})= -1.481+0.665 ln(CO_2)+\\ +0.334 ln(GDP)+0.134 ln(Population)                                            
\end{multline}

As observed from Eq. (\ref{eq4}), a 1\% change in CO$_2$ emissions results in 0.67\% increase in energy use. Moreover, a 1\% change in GDP per capita results in 0.33\% increase in energy use, while 1\% increase in population results in 0.13\% increase in energy use. All predictors in the model are statistically significant and impact on the energy use by increasing it. 

\begin{figure}[htbp]
\begin{center}
\includegraphics[height=7.2cm, width=7.2cm, keepaspectratio]{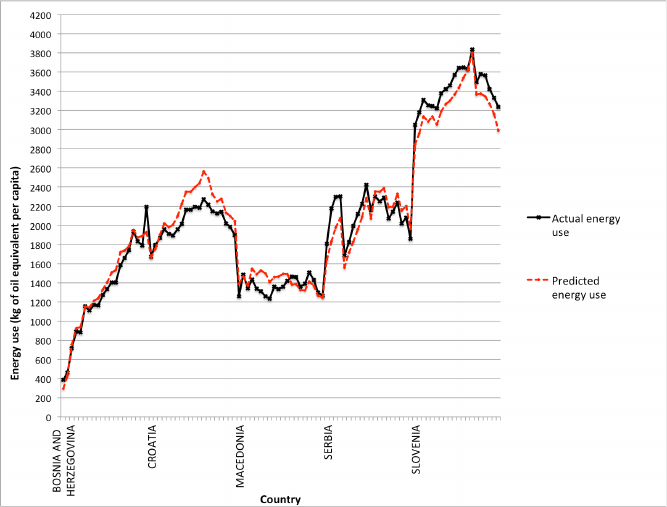}
\caption{Actual and predicted values of energy use (multiple linear regression method)}
\label{fig5}
\end{center}
\end{figure}

The regression equation was then applied to present and past values in order to compare the correctness of the regression model. Fig. \ref{fig5} shows the actual and predicted energy use. As observed, the predicted values are very similar to the actual values, with some deviations.

\subsection{Multilayer perceptron model}
Before employing a neural network model, the number of samples belonging to the training, testing and holdout sets was specified to be 70\% for training, 18\% for testing and 12\% for holdout. The scaled conjugate gradient was used as an optimisation algorithm. 

\begin{figure}[htbp]
\begin{center}
\includegraphics[height=7.2cm, width=7.2cm, keepaspectratio]{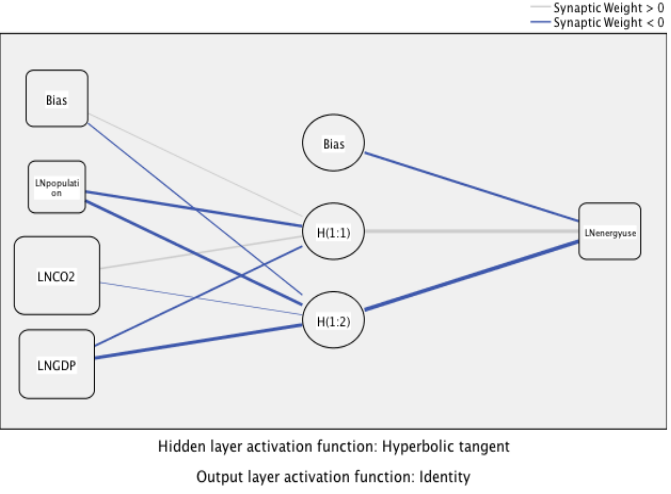}
\caption{Schematic representation of the multilayer perceptron model}
\label{fig6}
\end{center}
\end{figure}

There were two units in the hidden layer, and the schematic representation of the neural network model is represented in Fig. \ref{fig6}. The diagram shows 3 input nodes, 2 hidden nodes and one output node representing the energy use. 

Table \ref{tab3} presents the results after applying the multilayer perceptron network to the sample. As the output layer consists of a scale dependent variable, a sum of squares error is presented here. The sum of squares error is the error that the neural network tries to minimise during the training phase. The used stopping rule is one consecutive step with no decrease in the error. The relative errors for training, testing and holdout are very similar (0.013, 0.026 and 0.020, respectively), hence it can be assumed that the model is not overtrained. The computation of the errors was based on the testing set. 
\begin{table}[ht]
\caption{Errors of the multilayer perceptron model}
\begin{center}
\begin{tabular}{cc}
\includegraphics[height=6cm, width=6cm, keepaspectratio]{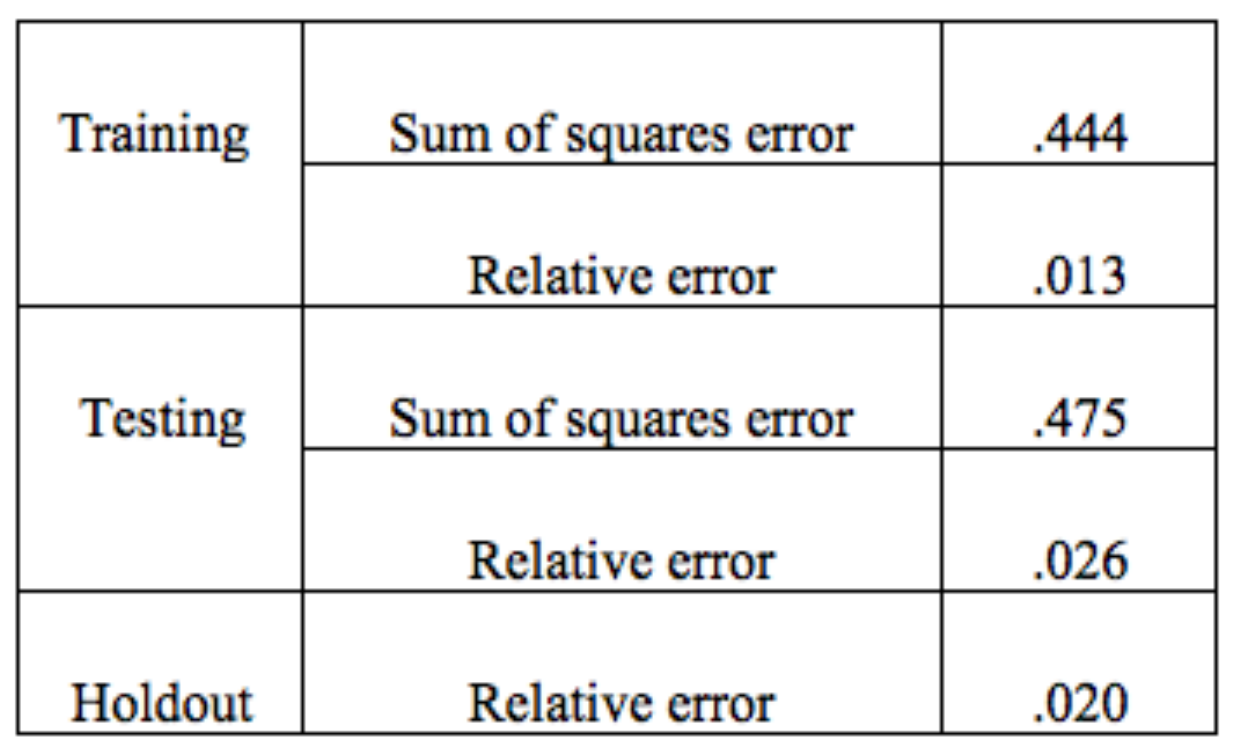}
\end{tabular}
\end{center}
\label{tab3}
\end{table}%  

\begin{figure}[ht]
\begin{center}
\includegraphics[height=7.2cm, width=7.2cm, keepaspectratio]{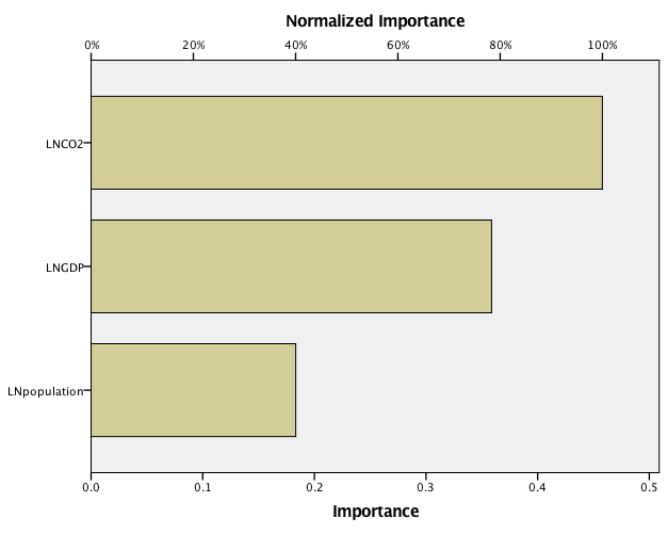}
\caption{Sensitivity analysis}
\label{fig7}
\end{center}
\end{figure}

It can be observed that the importance of the independent variable, i.e. how much the model predicted the value, changes for different values of the independent variables. A sensitivity analysis that calculates the importance of each predictor variable has been computed (see Fig. \ref{fig7}), and shows that the most important factors are the CO$_2$ emissions (in logarithmic format), followed by the GDP and Population. 

A comparison between actual values and values predicted by the multilayer perceptron model was performed. It can be observed that the deviations from the actual values were relatively small, and that in general this model predicts the energy consumption better than the regression model. This is presented in Fig. \ref{fig8}:
 \begin{figure}[ht]
\begin{center}
\includegraphics[height=7.2cm, width=7.2cm, keepaspectratio]{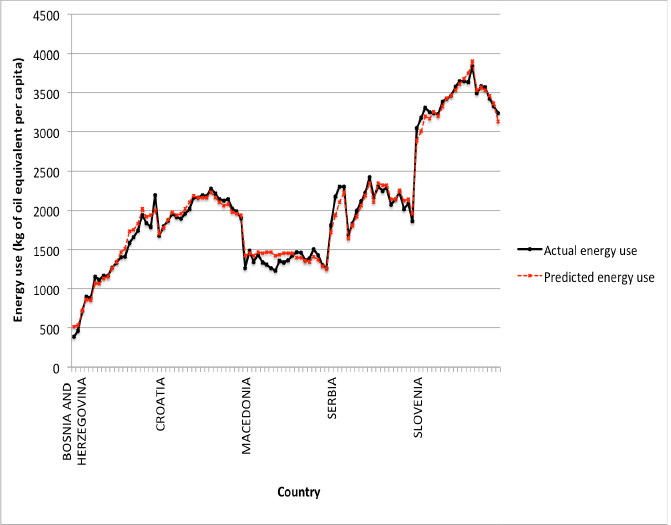}
\caption{Actual and predicted values of energy use (multilayer perceptron model)}
\label{fig8}
\end{center}
\end{figure}

Lastly, the RMSE and MAPE for both models were calculated. The RMSE is calculated as follows:
\begin{equation}
RMSE=\sqrt{\frac{\sum_{i=1}^n(O_i-P_i)^2}{n}}
\end{equation}

where $O_i$ is the $i$-th observed value, $P_i$ is the $i$-th predicted value, and $n$ is the number of samples in the dataset.
 
MAPE is calculated as follows:
\begin{equation}
MAPE=\frac{\sum_{i=1}^n|\frac{O_i-P_i}{O_i}|)}{n}\times100
\end{equation}

Both of these measures were compared in Table \ref{tab4}. Based on the calculations, it can be concluded that the multilayer perceptron model generally performs much better than the regression model, with lower values of both RMSE and MAPE. 
\begin{table}[ht]
\caption{Comparison of RMSE and MAPE for both models}
\begin{center}
\begin{tabular}{cc}
\includegraphics[height=7cm, width=7cm, keepaspectratio]{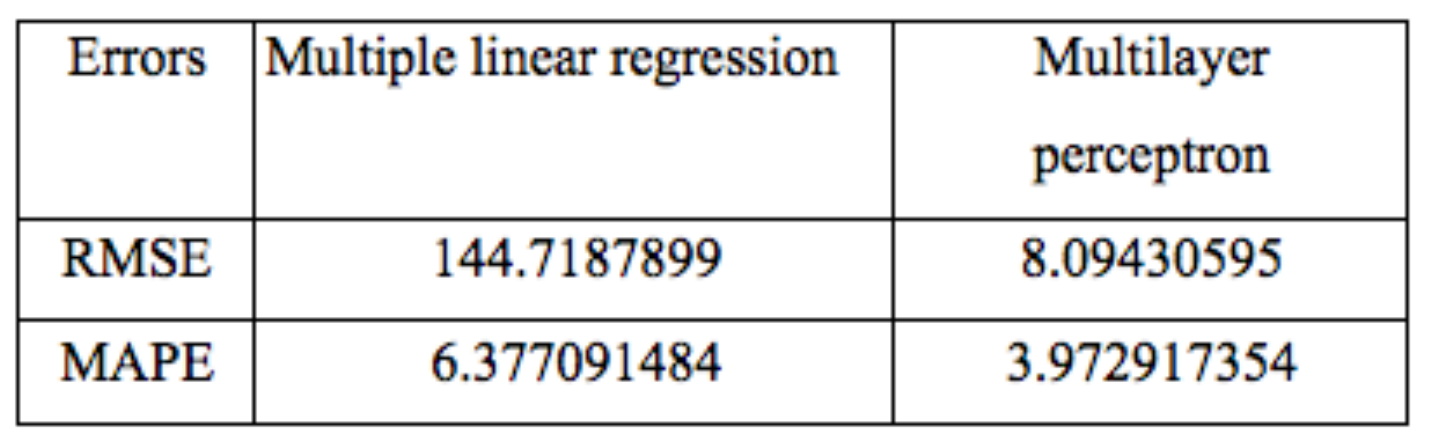}
\end{tabular}
\end{center}
\label{tab4}
\end{table}%  

\section{Conclusion}
The multiple regression model and the multilayer perceptron neural network model were developed to predict the energy use, using population, GDP and CO$_2$ emissions indicators. Both models performed relatively well, with the multi-layer perceptron model performing the best, with very small deviations from the actual values and lower values of RMSE and MAPE. 

The results indicate that the CO$_2$ emissions mostly influence the energy use, followed by the GDP indicator, while the population number influences the energy use relatively less. The neural network model performed much better compared to the multiple linear regression model. As observed, both models predicted the energy use more accurately for Bosnia and Herzegovina, while the neural network model performed with minimal deviations for all countries, except for Macedonia, where the deviations from the actual values were slightly bigger. Moreover, the CO$_2$ emissions are generally the best fitted predictors of energy use in the Balkan countries. The future research will involve developing a forecasting model based on different demographic and quality-of-life factors, with a focus on Serbia. 

\begin{center}
{\bf Acknowledgement}
\end{center}
This work was supported by the Mathematical Institute of the Serbian Academy of Sciences and Arts (Project III44006).

\end{document}